# FrustumFusionNets: A Three-Dimensional Object Detection Network Based on Tractor Road Scene

Lili Yang [a, b], Mengshuai Chang [a, b], Xiao Guo [a, b], Yuxin Feng [a], Yiwen Mei [a], Caicong Wu [a, b, *]

a. College of Information and Electrical Engineering, China Agricultural University, Beijing 100083, China
b. Key Laboratory of Agricultural Machinery Monitoring and Big Data Applications, Ministry of Agriculture and Rural Affairs, Beijing 100083, China

**Abstract:** To address the issues of the existing frustum-based methods' underutilization of image information in road three-dimensional object detection as well as the lack of research on agricultural scenes, we constructed an object detection dataset using an 80-line Light Detection And Ranging (LiDAR) and a camera in a complex tractor road scene and proposed a new network called FrustumFusionNets (FFNets). Initially, we utilize the results of image-based two-dimensional object detection to narrow down the search region in the three-dimensional space of the point cloud. Next, we introduce a Gaussian mask to enhance the point cloud information. Then, we extract the features from the frustum point cloud and the crop image using the point cloud feature extraction pipeline and the image feature extraction pipeline, respectively. Finally, we concatenate and fuse the data features from both modalities to achieve three-dimensional object detection. Experiments demonstrate that on the constructed test set of tractor road data, the FrustumFusionNetv2 achieves 82.28% and 95.68% accuracy in the three-dimensional object detection of the two main road objects, cars and people, respectively. This performance is 1.83% and 2.33% better than the original model. It offers a hybrid fusion-based multi-object, high-precision, real-time three-dimensional object detection technique for unmanned agricultural machines in tractor road scenarios. On the Karlsruhe Institute of Technology and Toyota Technological Institute (KITTI) Benchmark Suite validation set, the FrustumFusionNetv2 also demonstrates significant superiority in detecting road pedestrian objects compared with other frustum-based three-dimensional object detection methods.

Keywords: tractor road; multi-source sensors; frustum; hybrid fusion; three-dimensional object detection; neural networks;

## 1. Introduction

With the development of modern agriculture, unmanned farms have become an important part of intelligent agriculture, and intelligent farm machinery is the material support of unmanned farms(Luo et al., 2021). In the realm of agriculture, tractor roads present a complex scene for unmanned farm machines due to the absence of clear traffic signs and the diversity of objects present. Accurately identifying obstacles on these roads is essential for farm machinery's intelligent operation.

In the field of 3D object detection, numerous methods based on multi-source sensor data fusion have been proposed, with the most extensive research focusing on the fusion of RGB camera and LiDAR(Mao et al., 2023). Currently, the fusion methods of camera and LiDAR are mainly classified into three categories: early-fusion methods(Paigwar et al., 2021a; Qi et al., 2018a; Shin et al., 2019; Vora et al., 2020; Z. Wang & Jia, 2019a), which aim to merge the knowledge of the image into the point cloud, and use the rich contextual information of the image to assist in the enhancement of 3D detection of the point cloud; and intermediate-fusion methods(Bai et al., 2022; Chen et al., 2017a, 2023; Ku et al., 2018; Li, Qi, et al., 2022; Yoo et al., 2020), which aim to fuse the previously extracted features of the point cloud and the image in the stage of backbone network, proposal generation, or ROI refinement in order to complete the final detection; Late-fusion methods(Pang et al., 2020, 2022) are aimed at post-processing the outputs of the point cloud-based 3D object detector and the image-based 2D detector to produce more accurate detection results. Some of the early-fusion methods(Shin et al., 2019; Vora et al.,

2020) retain more information about the original data, but the huge amount of data requires high computational power. The intermediate-fusion methods simultaneously simplify data processing and extract valuable feature information from various modalities. However, various feature extraction methods significantly influence the fusion effect and are highly sensitive to feature selection. Moreover, late-fusion methods offer a significant degree of flexibility and scalability. However, there is a high loss of detail in the data.

Most of the existing methods are primarily designed for urban scenes, and research focused on agricultural environments is still in its early stages. In the rural tractor road scene, the object has more distinct class characteristics than the urban road. At the same time, due to the lack of road neatness on the tractor road and the occlusion formed by trees and debris along the roadside, these complex environmental factors significantly enhance the difficulty of 3D object detection. We propose a 3D object detection network called FrustumFusionNets (hereinafter referred to as FFNets) based on the fusion of RGB image and LiDAR point clouds in frustum space. Initially, we gather data synchronously on rural roads using a monocular camera and an 80-line LiDAR, creating an object detection dataset specifically for tractor roads. After that, we use a hybrid fusion method that combines the early-fusion and intermediate-fusion paradigms, first figuring out the object's 2D bounding box. We then restrict LiDAR's search space to the frustum, extract features from the point cloud and crop image, and concatenate and fuse these features, ultimately detecting the object's 3D attitude.

The contributions of this paper are mainly listed as follows:

(1) We construct an object detection dataset based on an 80-line LiDAR and RGB camera in a complex tractor road scene, which includes people in various postures, cars, cyclists, bicycles, trucks, freight tricycles, and a total of six categories of objects.

(2) We propose a 3D object detection network, FrustumFusionNets. This network introduces a Gaussian mask mechanism and further fuses the point cloud features with the image features in the detection stage, which improves the 3D object detection accuracy through hybrid fusion.

(3) On the constructed tractor road data, our method significantly improves the 3D object detection accuracy, distance error, and orientation error of six categories of objects compared with the original model. Additionally, our method outperforms other frustum-based 3D object detection methods in detecting the pedestrian category on the KITTI validation set.

## 2. Related work

In recent years, 3D object detection algorithms based on deep learning have achieved remarkable results. And because of the excellent complementary performance of the camera and LiDAR, the fusion method based on the camera and LiDAR has higher detection performance compared with other methods.

### 2.1 End-to-end 3D object detection network

The ContFuse (Liang et al., 2018) significantly improves the accuracy of 3D object detection by mapping image features to the bird's-eye view (BEV) space and combining it with the two-stream network architecture to achieve deep continuous fusion of point clouds and image features at the multi-scale level. Aiming at the unavoidable information loss problem during image projection, EPNet (Huang et al., 2020) proposes a point-based correspondence establishment method to adaptively evaluate the importance of image semantic features. On this basis, EPNet++ (Liu et al., 2022) further innovatively introduces the CB-Fusion module to generate more discriminative and comprehensive feature representations. The BEVFusion (Liu et al., 2023), on the other hand, effectively mitigates the dependence on image depth estimation by projecting fine-grained image features into the BEV space and fusing them with LiDAR features. In addition, some research projects adopt knowledge distillation techniques to achieve the fusion of multimodal information. For example, LIGA-Stereo (Guo et al., 2021) transfers geometrically-aware representations of point clouds to stereo images through knowledge distillation to facilitate information interaction. The UVTR (Li, Chen, et al., 2022) approach further characterizes images and point clouds in a unified manner while supporting cross-modal fusion and knowledge migration, enabling effective distillation from multimodal or continuous frames to a single input. Although these end-to-end fusion methods simplify the overall processing flow to some extent, they also bring about a reduction in model interpretability, creating challenges for model debugging and optimization. Additionally, these methods often require substantial data support for training, which may limit their effectiveness in datasets with a limited number of samples.

**2.2 3D object detection network based on frustum**

In the field of 3D object detection, in addition to end-to-end methods, multi-stage strategies have been widely adopted, of which the typical one is based on the frustum. FrustumPointNets (hereinafter referred to as FPointNets) (Qi et al., 2018b) adopts a two-stage approach. First, the 2D object detector is used to generate the 2D bounding box of the object, and then these 2D boxes are projected to form the frustum, which is applied to the 3D point cloud to narrow the search space and further detect the object in the point cloud. This is the first method to propose the frustum paradigm. In the same year, PointFusion (Xu et al., 2018) was also based on the frustum. In addition to using the detection results of the image to limit the search scope of the point cloud, it further integrated image features into the processing of the frustum point cloud. F-ConvNet (Z. Wang & Jia, 2019b) is further optimized on the basis of the FPointNets to extract more fine-grained features by subdividing the frustum region into smaller units. Frustum-PointPillars (Paigwar et al., 2021b), on the other hand, addresses the problem of independent processing of objects in previous approaches and proposes a unified framework for scene understanding, which significantly reduces the computation time. Frustum-based methods effectively reduce the complexity of the algorithms as well as the demand for computational resources and are highly flexible since they can be built on top of any state-of-the-art 2D detector. However, such methods tend to rely mainly on point cloud information in the 3D object detection stage and do not utilize image information sufficiently. Although PointFusion attempts to fuse RGB image and point cloud information in the second stage, its utilization of image and point cloud features is still insufficient, and its performance does not surpass that of FPointNets, a method that utilizes only point cloud information.

## 3. Materials and methods

### 3.1 Data set production

As shown in Fig. 1, the DF2204 tractor mounts the data acquisition platform, which includes the point cloud information acquisition module, image information acquisition module, and data storage and processing module. The point cloud information acquisition module is the RS-Ruby Lite mechanical 80-line LiDAR, installed on the crossbar at 1.33 m from the ground, with an acquisition frequency of 10 Hz, a vertical field of view of 40° (-25°~+15°), a vertical resolution of 0.1°, and a horizontal resolution of 0.2°. The image information acquisition module is the MV-CS050-10GMGC-PRO industrial camera with an MVL-MF0828M-8MP fixed-focus lens that is mounted on a crossbar 1.03 m above the ground and is activated by LiDAR. It has a resolution of 1224x1024, a horizontal field of view of 54.97°, and a vertical field of view of 47.06°. The acquisition frequency is 10 Hz. The data storage and processing module is a Nuvo-810GC type industrial computer, installed in the tractor cab, and receives the data collected by the camera and the LiDAR sensor via Ethernet protocol.

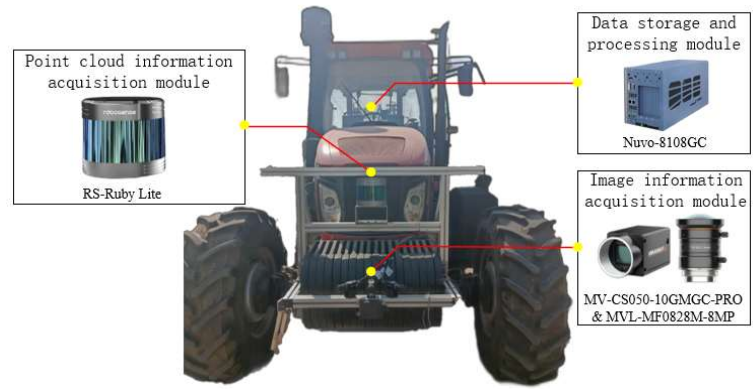

Figure 1 Data Acquisition Platform

In March 2024, we conducted the data collection in Henanzhai Town, Miyun District, Beijing. Given the local environment's complexity, the tractor collected data at a driving speed of 5 km/h. The data collection period spanned approximately one hour, during which we selected about 20 minutes of driving section data that contained richer object information. Given the slow driving speed of the tractor, we constructed the dataset by capturing one key frame for every 10 frames, ultimately obtaining a total of 1,295 valid frames. We set the region of interest (ROI) to be the overlapping range of the LiDAR and the camera's field of view and annotated the 3D bounding box for the six categories of dynamic obstacles within the ROI, which include cars, people, cyclists, bicycles, trucks, and freight tricycles, according to the KITTI(Geiger et al., 2013) standard. We utilized the camera and LiDAR's external parameters to project the 3D annotation truth values onto the image, thereby generating the 2D annotation truth values. Consequently, we constructed a multimodal 3D object detection dataset from the tractor road scene. We then compared this dataset with the annotated pedestrian categories found in the KITTI dataset. Our dataset features people in various postures, including sitting, squatting, tilting, and leaning. Table 1 shows the composition of the dataset, with the two main categories of "cars" and "people" accounting for a larger proportion of the data.

Table 1 Data set object statistics

| Category | Quantity | Avg. length (m) | Avg. width (m) | Avg. height (m) |
| --- | --- | --- | --- | --- |
| Cars | 1354 | 4.47 | 1.98 | 1.64 |
| People | 1246 | 0.69 | 0.75 | 1.57 |
| Cyclists | 204 | 1.87 | 1.00 | 1.64 |
| Bicycles | 141 | 1.70 | 1.22 | 1.13 |
| Trucks | 87 | 4.66 | 1.99 | 1.85 |
| Freight Tricycles | 393 | 2.36 | 2.36 | 1.12 |

**3.2 Method**

**3.2.1 Data pre-processing**

First, we divide the dataset into a training set, a validation set, and a test set in a ratio of 70%, 15%, and 15%, with 906 frames of the training set, 194 frames of the validation set, and 195 frames of the test set.

To prevent errors from accumulating during training, we directly utilize the 2D bounding box ground truth values for the training and validation sets to generate the corresponding frustum point clouds and crop image data. Set $P \in \mathrm{R}^3$ as a point in the point cloud space and transform it into its projection on the image plane $\bar{P} \in \mathrm{R}^2$ using the camera projection matrix $T: \mathrm{R}^3 \rightarrow \mathrm{R}^2$. Next, we remove the projected point cloud outside the 2D bounding box and extract the point cloud data from the object frustum. Following a specific cropping strategy, we generate the object crop image based on the object's 2D bounding box.

Meanwhile, for the training set, given that real 2D detection is generally unable to reach the annotated standard, we use perturbations of the 2D ground truth bounding box to simulate the real detection results and augment the data with multiple perturbations to increase the algorithm's robustness (Eq. 1).

$$\begin{cases} x' = x + w * shift\_ratio * \alpha \\ y' = y + h * shift\_ratio * \alpha \\ w' = w + w * shift\_ratio * \alpha \\ h' = h + h * shift\_ratio * \alpha \end{cases} \quad (1)$$

Where $x$, $y$, $w$, and $h$ are the center coordinates of the annotated 2D box and the width and height values, $\alpha$ is a random number $\in [-1,1]$, $shift\_ratio$ is the perturbation rate parameter, and the value is 0.1, which represents the random movement of the center point of the annotated 2D box and the random scaling of the width and height. Using the perturbation of the 2D bounding box to regenerate objects' frustum point cloud and crop image as training data, we perturb each object's 2D box five times, resulting in a total of 11,980 objects that match the frustum point cloud data and the crop image as training data. The original training set had 2396 objects. We use Yolov7 for preprocessing the test set. Firstly, it pre-trains Yolov7(C.-Y. Wang et al., 2023) on the COCO dataset, then fine-tunes it on the tractor road image dataset. Finally, the frustum point cloud and crop image data are generated using the object 2D bounding box created by the fine-tuned Yolov7.

**3.2.2 Frustum fusion networks**

We construct a 3D object detection network called FFNets based on the hybrid fusion paradigm, using FPointNets as its foundation.

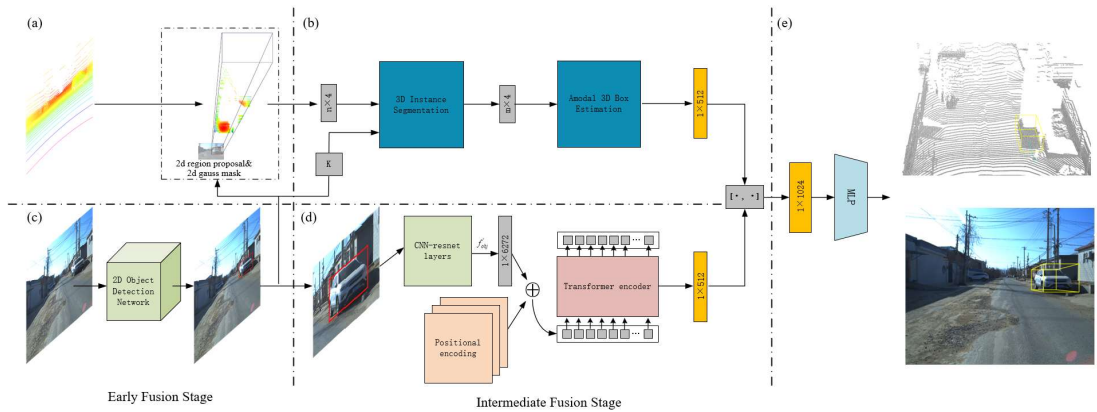

Fig. 2 FFNets architecture

Fig. 2 shows the architecture of our model. The original FPointNets model consists of four modules: (a), (b), (c), and (e). For modules (b), FPointNets provides two architectures based on PoiNet and PointNet++, which correspond to the versions FPointNetv1 and FPointNetv2. In this paper, we introduce the Gaussian mask mechanism in the box part of module (a) to enhance the point cloud features, and we add the image 3D object feature extraction module (d). Finally, we combine the five modules (a), (b), (c), (d), and (e) to form the FFNets. Module (c) can be substituted for any 2D detection network. FFNets also includes two versions, FFNetv1 and FFNetv2, which correspond to the two architectures of module (b). In Fig. 2, modules (a) and (b) make up the point cloud feature extraction pipeline, and the modules (c) and (d) make up the image feature extraction pipeline. Modules (a) and (c) form the early-fusion, while the modules (b) and (d) form the intermediate-fusion. The FFNets take modal data from both the 3D point cloud and the RGB image and, by inference, output the object's 3D information.

**3.2.2.1 Point cloud feature extraction pipeline**

Modules (a) and (b) in Fig. 2 illustrate the point cloud feature extraction pipeline in the FFNets. We detect 2D objects by using regular rectangles that contain the minimum of the object, but the frustum point cloud obtained through it typically doesn't precisely describe the object. In addition to the object itself, the points in the space frequently contain foreground and background noise. In real-world scenes, the diversity of human postures and grouping, as well as the variety of vehicle shapes and susceptibility to occlusion, complicate object detection in such scenarios. Therefore, distinguishing between foreground and background noise becomes very difficult. The same problem exists in the scene of tractor roads.

We try to solve the above problem by using a 2D Gaussian mask to describe the centrality of the points in the frustum point cloud (Eq. 2), based on the fact that the object is more likely to occupy the center region of the 2D box.

$$\xi(x, y) = \exp\left(-\frac{(x-x_0)^2}{2w^2} - \frac{(y-y_0)^2}{2h^2}\right) \quad (2)$$

Where $x$ and $y$ represent the projected coordinates of the point cloud on the image plane. $x_0$ and $y_0$ represent the center points of the 2D bounding box, while $w$ and $h$ represent its width and height. We add the image Gaussian mask values to the point cloud based on where the point cloud projection is on the image plane. This enables the expansion of the frustum point cloud features to four dimensions and the conversion of the point cloud channel features to $D = (x、y、z、\xi)$. According to Eq. 2, a point's proximity to the object's box center correlates with a higher mask value, while a point's distance from the center results in a smaller mask value. Since the object is more likely to occupy the center region, adding one-dimensional Gaussian mask features as weights to the point cloud allows for a rough distinction between foreground and background noise without significantly increasing the complexity (see module (a)).

Next, we sample $n = 1024$ points from the frustum point cloud, attaching Gaussian mask features to them. We then input these points into the 3D instance segmentation module, along with the object category code $k$ that we identified from the image. In the segmented object point cloud, $m = 512$ object points are chosen, and the object points' center of mass is used as the initial center point. We then panned the point cloud to the center point and fed it into the amodal 3D box estimation module, which extracted it into a $1 \times 512$ point cloud feature vector (see module (b)).

**3.2.2.2 Image feature extraction pipeline**

Modules (c) and (d) in Fig. 2 illustrate the image feature extraction pipeline in the FFNets. The pipeline starts by using a 2D detection network to get the object in the image's 2D detection results, as shown in module (c). Module (d) illustrates how we use the 2D detection results to obtain the crop image, which we then input into the image 3D feature extraction module that employs the transformer encoder(Carion et al., 2020; Dosovitskiy et al., 2020) architecture. Because the point cloud is limited to the range of frustum space according to the 2D detection results of the image, cropping the output of module (c) is necessary to focus on the object and eliminate redundant information. While considering the transformer's advanced global modeling capabilities(Carion et al., 2020), the background information is also crucial for feature extraction. Therefore, we apply cropping along the object's long side to encompass certain background information (Eq. 3).

$$L = \max(w, h) * \alpha \quad (3)$$

Where $w$ and $h$ are the width and height of the object's 2D bounding box, $\alpha = 1.5$ is the expansion scale parameter, and $L$ is the crop image's side length.

The initial image, $x_{obj} \in R^{3 \times L \times L}$, is a crop image with an attached background (with 3 color channels). After resizing the crop image to a size of $224 \times 224$ and feeding it into a conventional CNN backbone consisting of 4 layers of resnet-layers(He et al., 2016), we generate a low-resolution activation map $f_{obj} \in R^{C \times l' \times l'}$, C=512, $l'$=7. Then, by $1 \times 1$ convolution, the activation map $f_{obj}$ with high-dimensional features is reduced to a smaller dimension to create a new feature map $f'_{obj} \in R^{d \times l' \times l'}$, $d = 128$. In order to adapt to the serialized input of the transformer encoder, we stretch the two-dimensional feature map into a one-dimensional vector, which further yields a $1 \times 6272$ feature vector. After that, the feature vector is input into the transformer encoder module. Each encoder layer has a standard structure with an 8-head self-attention module and a feed-forward networks (FFN). We add positional encodings to the feature vector before sending it to the transformer encoder module. After three identical encoder layers, we can get a $1 \times 512$ crop image feature with background data.

### 3.2.2.3 Hybrid Fusion

The FFNets use hybrid fusion to achieve 3D object detection. Firstly, in the early stage, the network generates a frustum point cloud based on the detection results of the 2D image and crops the object in the image to generate a crop image. The network computes the Gaussian mask corresponding to each point, based on their 2D coordinates projected onto the image plane, to enhance the point cloud's features. This interaction between the point cloud and the image is known as the early-fusion. In the intermediate stage, two pipelines are used to get the features from the point cloud and RGB image data and generate two 1x512 feature vectors. We then concatenate the two feature vectors to generate a 1x1024 high-dimensional fused feature vector. This interaction between the point cloud and the image is referred to as intermediate-fusion. Subsequently, we feed the fused high-dimensional feature vectors into a multilayer perceptron (MLP) to obtain the 3D prediction results of the object, which include its location, size, and orientation, as illustrated in module (e) of Fig. 2. Given that the size differences among the same class of objects in the tractor road scene will not be significant, we use the mean values of the length, width, and height of the object in Table 1 as a priori information and further predict the residuals of the length, width, and height of the object to obtain the 3D dimensions of the object. For the orientation prediction, we employ a hybrid method of classification and regression(Mousavian et al., 2017), classifying the 360-degree orientations into 12 classes of 30 degrees each. After classification, we further regress the angle within the category to obtain an accurate orientation prediction.

## 4. Experimental results and discussion

Experiments to evaluate the performance of FFNets were conducted on both the self-constructed tractor road dataset and the KITTI dataset.

## 4.1 Model Evaluation on the Tractor Road Dataset

The model was implemented based on TensorFlow 1.4, and the models before and after the improvement were trained on the constructed tractor road dataset, respectively, with the training platform of 3090-24G. After 200 iterations of training, the model's accuracy (IoU > 0.7) and inference time on the validation set are shown in Table 2.

Table 2 Comparison of models on val set

| Model | 3D box estimation accuracy % (IoU>0.7) | Inference speed ms |
|---|---|---|
| FPointNetv1 | 87.29 | 14.67 |
| FFNetv1 | 91.45 | 19.32 |

As seen in Table 2, the FFNetv1 model, with the introduction of the Gaussian masking mechanism and the image 3D object feature extraction module, improves the average accuracy of the 3D box estimation accuracy on the validation set by 4.16% compared to FPointNetv1. Meanwhile, the model's average inference time rose by 4.65 ms. We conducted further ablation studies on the validation set of the tractor road, and Table 3 shows the comparison results.

Table 3 3D object detection accuracy for the tractor road val set

| Model | Gaussian mask | Image 3D feature extraction module | | accuracy (IoU>0.7) % |
| | | Crop (excluding background) | Crop (including background) | |
|---|---|---|---|---|
| FPointNetv1 | - | - | - | 87.29 |
| FFNetv1 | √ | - | - | 89.16 |
| FFNetv1 | √ | √ | - | 90.21 |
| FFNetv1 | √ | - | √ | 91.45 |

Note: √ indicates that the relevant policy is used - indicates that the relevant policy is not used

Table 3 shows that in the validation set, the introduction of the Gaussian mask mechanism improves the 3D detection accuracy of the FFNetv1 model for the object by 1.87%, which proves that the Gaussian mask mechanism distinguishes the foreground from the background noise in the frustum point cloud to some extent. The addition of the image 3D feature extraction module enhances the detection accuracy of FFNetv1 by 1.05% when using a crop image without background information and by 2.29% when using a crop image with background information. This indicates that the rich contextual information of the image significantly aids in the detection of 3D objects. Additionally, the global modeling capability of the transformer architecture enables the 3D image feature extraction pipeline to more effectively extract 3D information from the crop image with specific background information, thereby improving the final detection results.

Based on FFNetv1, FFNetv2 is constructed by replacing the point cloud feature extraction module in Fig. 2(b) with the PointNet++ architecture. We compared the performance of FFNetv1, FFNetv2, and FPointNetv2 on the validation set, and the results are shown in Table 4.

Table 4 Comparison of models on the tractor road val set

| Model | 3D box estimation accuracy% (IoU>0.7) | Inference speed ms |
| --- | --- | --- |
| FPointNetv2 | 90.00 | 23.01 |
| FFNetv1 | 91.45 | 19.32 |
| FFNetv2 | 92.91 | 25.73 |

According to Table 4, FFNetv2 improves the 3D frame detection accuracy by 1.46% compared to FFNetv1, but the inference time increases by 6.41 ms. It can be seen that the improvement of the point cloud feature extraction module in the original model, which is also applicable to the model in this paper, can still significantly increase the accuracy rate with a limited increase in the elapsed time. It can also be noticed that FFNetv1 increases the accuracy by 1.45% compared to FPointNetv2 of the original model, while the time consumed decreases by 3.69 ms, which further proves the effectiveness and superiority of the Gaussian masking mechanism with the image 3D feature extraction module.

Table 5 shows the 3D detection AP for each network on the test set.

Table 5 AP (%) on the tractor road val set for 3D object detection.

| Model | Cars (IoU ≥ 0.7) | People (IoU ≥ 0.5) | Cyclists (IoU ≥ 0.5) | Bicycles (IoU ≥ 0.5) | Trucks (IoU ≥ 0.7) | Freight Tricycles (IoU ≥ 0.6) |
|---|---|---|---|---|---|---|
| FPointNetv1 | 78.72 | 92.09 | 74.54 | 91.90 | 83.29 | 96.16 |
| FPointNetv2 | 80.45 | 93.35 | 76.80 | 92.99 | 90.19 | 97.55 |
| FFNetv1 | 81.94 | 94.04 | 76.06 | 93.13 | 89.29 | 96.88 |
| FFNetv2 | **82.28** | **95.68** | **78.77** | **93.36** | **91.19** | **97.76** |

The results in Table 5 show that we propose that FFNetv2 outperforms FPointNetv2, the original model's best method, in all detection categories. It's better by 1.83% for cars and 2.33% for people. Our proposed FFNetv1 improves FPointNetv1 by 3.22% and FPointNetv2 by 1.49% in car detection and 1.95% and 0.69% in people detection, respectively. Among all the models, the accuracy of cyclists detection is low, and the results of each model are similar on the detection of bicycles or freight tricycles, with large differences in the truck category. This may be due to the small number of these categories in the test set and the influence of some outliers. We further evaluated the object detection results of each category in terms of distance and orientation, using NuSences(Caesar et al., 2020) evaluation metrics, to mitigate the issue of data volume discrepancies and anomalies.

Table 6 Distance error (m) results on the tractor road test set

| Model | Cars | People | Cyclists | Bicycles | Trucks | Freight Tricycles |
|---|---|---|---|---|---|---|
| FPointNetv1 | 0.677 | 0.482 | 0.309 | 0.388 | 1.626 | 0.256 |
| FPointNetv2 | 0.588 | 0.468 | 0.245 | 0.265 | 1.354 | 0.288 |
| FFNetv1 | 0.501 | 0.334 | **0.120** | 0.254 | 1.102 | **0.054** |
| FFNetv2 | **0.389** | **0.246** | **0.120** | **0.184** | **0.961** | 0.060 |

Table 6 demonstrates that FFNetv1 outperforms the two versions of the original model in terms of distance error in all categories. FFNetv2 is optimal in object detection for all categories except freight tricycles, and it further reduces the distance errors for the two main categories of cars and people by 0.112 m and 0.088 m from FFNetv1, which is 0.199 m and 0.222 m less than the original model optimal method, FFNetv2. Additionally, we have also evaluated the orientation error in object detection, and the results are shown in Table 7.

Table 7 Orientation error (rad) results on the tractor road test set

| Model | Cars | People | Cyclists | Bicycles | Trucks | Freight Tricycles |
|---|---|---|---|---|---|---|
| FPointNetv1 | 0.213 | 0.185 | 0.171 | 0.241 | 0.219 | 0.207 |
| FPointNetv2 | 0.213 | 0.178 | 0.158 | 0.227 | 0.200 | 0.204 |
| FFNetv1 | 0.193 | 0.173 | 0.197 | 0.212 | 0.201 | 0.206 |
| FFNetv2 | **0.178** | **0.154** | **0.180** | **0.204** | **0.199** | **0.182** |

Table 7 demonstrates that FFNetv2 excels in all object detection categories, reducing orientation errors by 0.015 rad and 0.019 rad compared to FFNetv1, and by 0.035 rad compared to the original model optimal method, FPointNetv2, and 0.024 rad.

In summary, our proposed FFNetv2 has superior performance in terms of object detection accuracy, distance error, and orientation error on the tractor road dataset compared to the two models of FFNetv1 and FPointNets, and the visualization results are shown in Fig. 3.

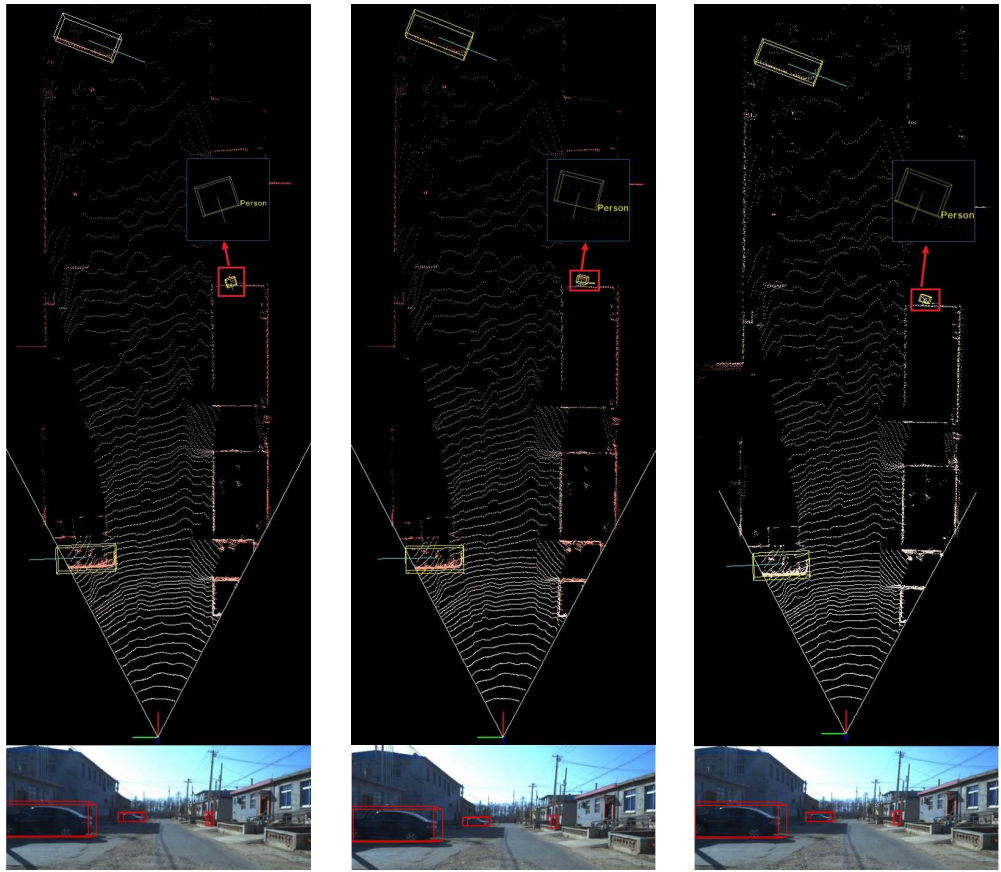
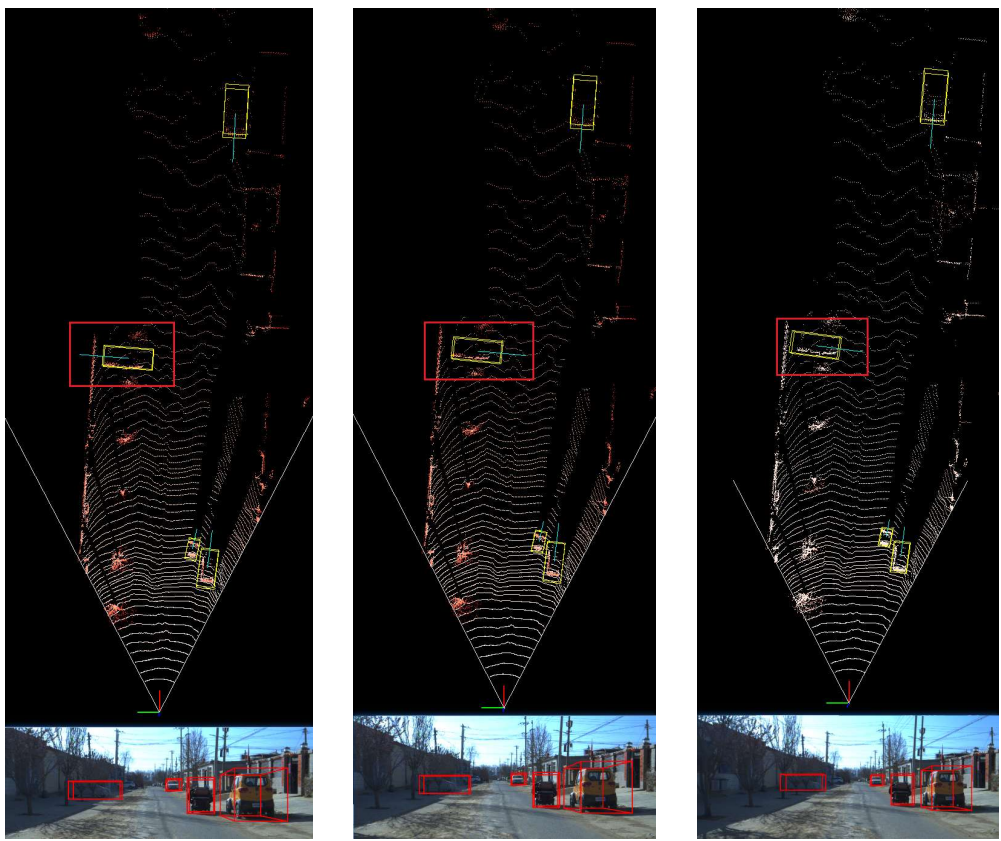

(a)                (b)                (c)

Fig. 3 Comparison of model detection visualization for the tractor road dataset

Fig. 3(a) shows the detection results of the original model FPointNetv2, Fig. 3(b) shows the detection results of the proposed model FFNetv2, and Fig. 3(c) shows the annotated true values. We can clearly see the superiority of FFNetv2 by comparing the objects annotated by the red box lines in the point cloud diagram. For the ground truth of the distance between the person and the farm machine annotated in Frame I, which is 36.426 m, the ground truth of the orientation is 1.875 rad. FFNetv2 detects the distance of 36.356 m with an error of 0.070 m, detects the orientation of 1.784 rad, and has an error of 0.091 rad, while FPointNetv2 detects the distance of 36.098 m with an error of 0.328 m, detects the orientation of 1.281 rad, and has an error of 0.594 rad. For the car orientation indicated in Frame II, FFNetv2 is superior to FPointNetv2. The ground truth of the indicated car orientation is 0.122 rad; FFNetv2 detects orientation at 0.074 rad with an error of 0.048 rad, while FPointNetv2 detects orientation at -3.107 rad with an error of 3.054 rad.

**4.2 Model evaluation on the KITTI dataset**

For a better comparison with other models, we further tested FFNetv2 on KITTI data. The KITTI dataset mainly contains 2D and 3D annotations of cars, pedestrians, and cyclists in urban driving scenes. Sensor configurations include a wide-angle camera and a Velodyne HDL-64E LiDAR. The official training set contains 7481 images. We divided the dataset into a training set and a validation set according to MV3D(Chen et al., 2017b), with each set accounting for about half of the entire dataset. The performance of the FFNetv2 model is tested on the validation set and compared with the original model-optimal network FPointNetv2 as a baseline, and the metrics on the validation set are shown in Tables 8 and 9.

Table 8 AP (%) on KITTI val set for BEV detection

| Model | Category | Easy | Mod | Hard |
| --- | --- | --- | --- | --- |
| FPointNetv2 | Cars | 88.16 | 84.02 | 76.44 |
| | Pedestrians | 72.38 | 66.39 | 59.57 |
| | Cyclists | 81.82 | 60.03 | 56.32 |
| | mean | 80.79 | 70.15 | 64.11 |
| FFNetv2 | Cars | 88.33 | 85.68 | 77.37 |
| | Pedestrians | 75.99 | 68.10 | 60.80 |
| | Cyclists | 80.88 | 62.65 | 58.31 |
| | mean | 81.73 | 72.14 | 65.49 |

Table 9 AP (%) on KITTI val set for 3D object detection.

| Model | Category | Easy | Mod | Hard |
| --- | --- | --- | --- | --- |
| FPointNetv2 | Cars | 83.76 | 70.92 | 63.65 |
| | Pedestrians | 70.00 | 61.32 | 53.59 |
| | Cyclists | 77.15 | 56.49 | 53.37 |
| | mean | 76.97 | 62.91 | 56.87 |
| FFNetv2 | Cars | 85.37 | 73.75 | 65.55 |
| | Pedestrians | 70.28 | 64.45 | 56.95 |
| | Cyclists | 79.39 | 60.53 | 56.46 |
| | mean | 78.35 | 66.24 | 59.65 |

As shown in Tables 8 and 9, FFNetv2 has improved in most metrics compared to the original model network, both in BEV detection accuracy and 3D detection accuracy. We also observe that our model significantly outperforms the original model in moderate and hard object detection; in Table 8, FFNetv2 achieves average improvements of 1.99% and 1.38% in moderate and hard object detection, respectively, compared to FPointNetv2, and in Table 9, FFNetv2 outperforms FPointNetv2 by an average of 3.33% and 2.78% in moderate and hard object detection, respectively. This indicates that FFNetv2 improves more than the original model when faced with occluded objects, and we further demonstrate this result in Fig. 4.

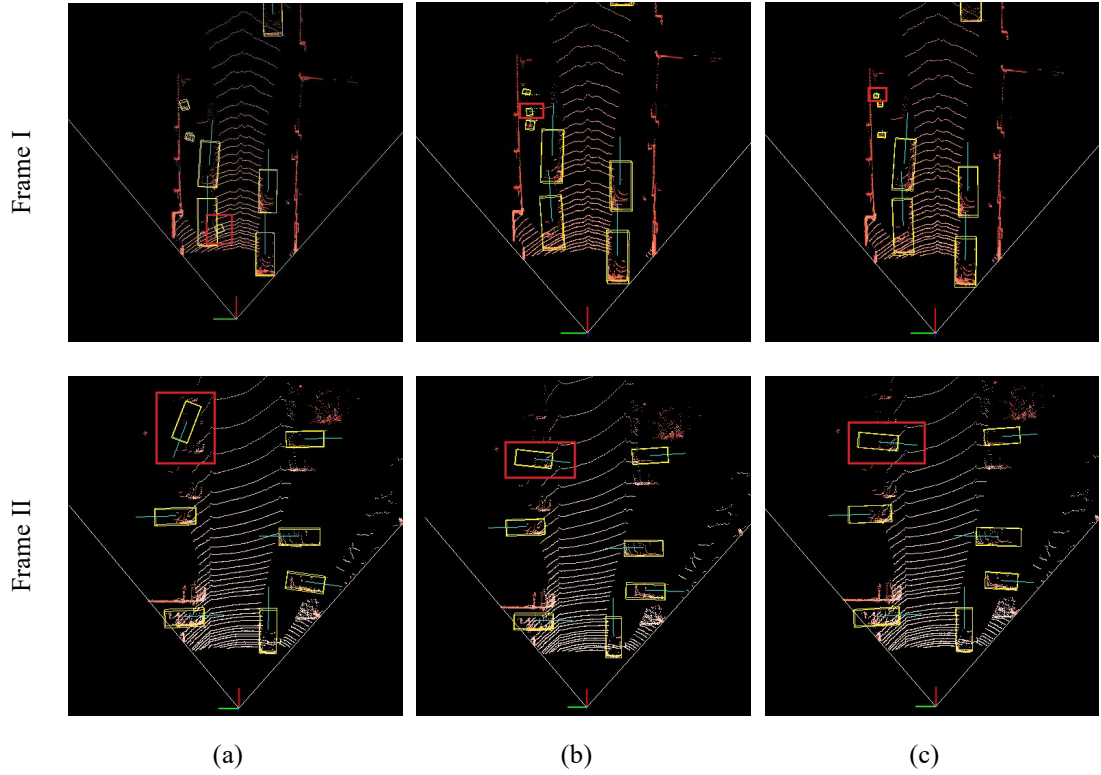

Fig. 4 Comparison of KITTI urban scene model detection visualization

Fig. 4(a) shows the detection results of the original model, FPointNetv2. Fig. 4(b) shows the detection results of our proposed model, FFNetv2. Fig. 4(c) shows the true values. For the distance of the pedestrians annotated in Frame I, the ground truth of the pedestrian's distance is 19.553 m, the detection distance of FFNetv2 is 17.432 m, and the error is 2.121 m. The FPointNetv2 detects a distance of 7.944 m, with an error of 11.609 m. The ground truth of the car orientation annotated in Frame II is 0.090 rad; the detection orientation of FFNetv2 is 0.107 rad with an error of 0.017 rad; and the detection orientation of FPointNetv2 is 1.931 rad with an error of 1.841 rad. We can significantly see the superiority of FFNetv2 for 3D object detection under occlusion.

We further compared FFNetv2 with the current representative frustum-based methods, FConvNet (Z. Wang & Jia, 2019a) and FPointpillars (Paigwar et al., 2021a), on the KITTI validation set. Table 10 shows the results of the BEV detection, while Table 11 shows the results of the 3D object detection.

Table 10 AP (%) on KITTI val set for BEV detection.

| Model | Cars | | | | Pedestrians | | | | Cyclists | | | |
|---|---|---|---|---|---|---|---|---|---|---|---|---|
| | Easy | Mod | Hard | Mean | Easy | Mod | Hard | Mean | Easy | Mod | Hard | Mean |
| FPointNetv2 | 88.16 | 84.02 | 76.44 | 82.87 | 72.38 | 66.39 | 59.57 | 66.11 | 81.82 | 60.03 | 56.32 | 66.06 |
| FConvNet | **90.42** | 88.99 | 86.88 | 88.76 | - | - | - | - | - | - | - | - |
| FPointpillars | 90.20 | **89.43** | **88.77** | **89.47** | 72.17 | 67.89 | **63.46** | 67.84 | **88.58** | **76.79** | **74.80** | **80.06** |
| FFNetv2 | 88.33 | 85.68 | 77.37 | 83.79 | **75.99** | **68.10** | 60.80 | **68.30** | 80.88 | 62.65 | 58.31 | 67.28 |

In terms of pedestrian BEV detection, our proposed FFNetv2 has the highest average accuracy of 68.30%, which is optimal in the simple pedestrian and medium pedestrian categories but lower than FPointpillars in the difficult category. With FPointpillars, BEV detection of cars and cyclists is better than FFNetv2. The evaluation of FConvNet did not include pedestrians and cyclists, and in the four model comparisons, it demonstrated only a slight advantage in the easy category of cars.

The 3D object detection is based on BEV detection with the addition of height dimension information, which is more difficult compared to the BEV detection, and the comparison of the detection effects of specific models is shown in Table 11.

Table 11 AP (%) on KITTI val set for 3D object detection.

| methodologies | Cars | | | | Pedestrians | | | | Cyclists | | | |
|---|---|---|---|---|---|---|---|---|---|---|---|---|
| | Easy | Mod | Hard | Mean | Easy | Mod | Hard | Mean | Easy | Mod | Hard | Mean |
| FPointNetv2 | 83.76 | 70.92 | 63.65 | 72.78 | 70.00 | 61.32 | 53.59 | 61.64 | 77.15 | 56.49 | 53.37 | 62.34 |
| FConvNet | **89.31** | 79.08 | 77.17 | 81.85 | - | - | - | - | - | - | - | - |
| FPointpillars | 88.90 | **79.28** | **78.07** | **82.08** | 66.11 | 61.89 | 56.91 | 61.64 | **87.54** | **72.78** | **66.07** | **75.46** |
| FFNetv2 | 85.37 | 73.75 | 65.55 | 74.89 | **70.28** | **64.45** | **56.95** | **63.89** | 79.39 | 60.53 | 56.46 | 65.46 |

In the 3D object detection results for pedestrians, our proposed FFNetv2 achieves optimal results in all three levels of 3D detection: easy, moderate, and hard, with an average accuracy of 63.89%. In the 3D object detection results for cars, FPointpillars has the best average accuracy, and in the 3D object detection results for cyclists, FPointpillars has the best results in each category. FConvNet only has a slight advantage in simple car category detection.

Comparing the BEV detection and 3D object detection results of FFNetv2 and FPointpillars in the KITTI validation set, there is a gap between FFNetv2 and FPointpillars in both cars category detection, with a bigger gap in 3D object detection; in cyclists detection, there is a bigger gap in BEV detection; and in pedestrians BEV detection, the FFNetv2 improves compared to FPointpillars, and more in 3D object detection; in particular, the accuracy of BEV detection for the difficult category of pedestrians is lower in FFNetv2 than in FPointpillars, but the accuracy of detection in 3D bounding box is higher in FFNetv2 than in FPointpillars, which illustrates that for small objects with occlusions, the introduction of a priori information about the object's length, width, and height may improve the height dimension recognition of FFNetv2.

## 5. Conclusions

Aiming at the problems of underutilization of image information by frustum-based object detection methods and less research work on 3D object detection in agricultural tractor road scenes, we created an object detection dataset specifically for agricultural scenes, and subsequently developed an 3D object detection network known as FFNets. This network can be based on any mature 2D object detection network to achieve real-time 3D object detection. Our major contributions are as follows:

(1) We collected 80-line LiDAR point cloud and RGB image data in the tractor road scene using the DF2204 tractor's data acquisition platform. We then annotated six categories of common objects in the tractor road scene and compared them with the KITTI dataset, which includes data from people in richer poses. Finally, we constructed a 3D object detection dataset for the tractor road scene.

(2) The point cloud data itself contains object location information, but the foreground and background noise around the object will reduce the 3D object detection accuracy. The image data itself contains object context information but lacks 3D location information. Increasing the background information of the image can help the network better understand the location of the object in the scene and thus improve the 3D detection accuracy. Our proposed FFNets take into account the characteristics of the two modalities of data for 3D object detection by introducing Gaussian masks to roughly distinguish the front background noise in the frustum point cloud and cropping the image to retain the object's background information, which effectively improves the detection effect. FFNetv1 achieves superior results in 3D object detection accuracy, distance error, and direction error on the constructed tractor road dataset for the two primary categories of people and cars. Further, on the basis of FFNetv1, FFNetv2 is proposed with even better performance. Compared with the original model optimal method, FPointNetv2, FFNetv2 improves the 3D frame accuracy by 1.83% and 2.33% on the two main categories of people and cars, respectively, and reduces the distance error by 0.199 m and 0.222 m and the direction error by 0.035 rad and 0.024 rad. The performance is optimized, while the average inference time of single frame data is about 26 ms, which can meet the real-time requirement of driverless agricultural machines traveling safely on the tractor road. On the KITTI validation set, FFNetv2 has a significant improvement over FPointNetv2 in terms of moderate- and hard-level object detection results. Compared with other methods in the frustum-based class, FFNetv2 performs excellently in detecting people compared with other models. Especially for occluded pedestrians, it is more advantageous in 3D detection compared to other models, which indicates that FFNetv2 has an advantage over other models in the detection of pedestrians, a smaller object.

(3) FFNetv2 is based on the frustum point cloud generated from 2D objects for point cloud feature extraction. Although this helps to achieve more accurate 3D detection, at the same time, the correlation between time consumption and the number of objects in the scene is higher in FFNetv2 as compared to the methods that process the whole scene data. For Frame II in Fig. 3, which contains four objects, the inference time of FFNetv2 is 20.91 ms, and if we reduce one car in the frame, the inference time is reduced to 17.05 ms, and the inference time is further reduced to 14.67 ms by reducing two cars. In the tractor road dataset validation set, FFNetv2's average inference time is 25.73 ms. This can meet the real-time 3D object detection needs of most tractor road scenes, but it still limits the model's use cases. In our future work, we will further improve this deficiency.

**DECLARATION OF COMPETING INTEREST**

The authors declare that they have no known competing financial interests or personal relationships that could have appeared to influence the work reported in this paper.

**ACKNOWLEDGMENT**

The research was financially supported by the National Key Research & Development Program of China (Grant/Contract number: 2023YFE0208200).